\documentclass[letterpaper, 10 pt, conference]{ieeeconf}  

\IEEEoverridecommandlockouts                              
\usepackage[ruled]{algorithm2e}
\usepackage{mathtools}
\usepackage{subcaption}
\usepackage{multirow}
\usepackage{multicol}
\usepackage[pagebackref=true,breaklinks=true,letterpaper=true,colorlinks,bookmarks=false]{hyperref}

\newcommand{\etal}{\textit{et al}. }

\title{\LARGE \bf
Weighted Average Precision: Adversarial Example Detection in the Visual Perception of Autonomous Vehicles
}

\author{
Yilan Li, Senem Velipasalar\\
Syracuse University\\
\{yli41, svelipas\}@syr.edu
}

\begin{document}

\maketitle
\thispagestyle{empty}
\pagestyle{empty}

\begin{abstract}
Recent works have shown that neural networks are vulnerable to carefully crafted adversarial examples (AE). By adding small perturbations to input images, AEs are able to make the victim model predicts incorrect outputs. Several research work in adversarial machine learning started to focus on the detection of AEs in autonomous driving. However, the existing studies either use preliminary assumption on outputs of detections or ignore the tracking system in the perception pipeline. 
In this paper, we firstly propose a novel distance metric for practical autonomous driving object detection outputs. Then, we bridge the gap between the current AE detection research and the real-world autonomous systems by providing a temporal detection algorithm, which takes the impact of tracking system into consideration. We preform evaluation on Berkeley Deep Drive (BDD) and CityScapes datasets to show how our approach outperforms existing single-frame-mAP based AE detections by increasing 17.76\% accuracy of performance.
\end{abstract}
\section{Introduction}
\label{sec:intro}

Significant progress in Machine Learning (ML) techniques like Deep Neural Networks (DNNs) recently has enabled the development of safety-critical ML systems like autonomous vehicles. Neural network based object detection models are widely implemented as an import part of the autonomous driving perception system. Recent results show that autonomous vehicles have become very efficient in practice and already driven millions of miles without any human interventions. Twenty US states including California, Texas, and New York have recently passed legislation to enable testing and deployment of autonomous vehicles. Since the control successors are highly depended on the outputs from the object detector and neural networks~\cite{ylu01}\cite{ylu02}\cite{ylu03}, the performance of object detection is significant to the security of autonomous driving. However, neural network based object detectors are shown to be vulnerable to adversarial examples by recent researches, inputs that are carefully crafted to fool the model~\cite{jia2019enhancing}\cite{lu2019enhancing}\cite{jia2019tracking}\cite{ylu05Jia2020Fooling}. Even worse, these perturbed images, if carefully crafted, can stay adversary when taken as input from the physical world using a camera. With development of neural networks, AEs are still threating the robustness\cite{ylu15}\cite{zzy01}. In order to solve this problem, many researches are focusing on improving the robustness of neural networks in order to totally defend the adversarial attacks\cite{zzy02}\cite{zzy03}\cite{zzy04}.  Unfortunately, until now, attacking technology is more advanced to that of defending. Attacking methods are able to fool the system with high success rate. 
In this situation, some researches turn to focus on inventing adversarial examples detection methods, which rise alarms when adversarial attacks are detected during the runtime. One the successful detecting method is perturbation detection. Perturbation based detecting methods hypothesis that the generated adversarial images, different to benign samples, are unstable to perturbations. Adding perturbation to adversarial examples makes a neural network model predicts very different results, while adding perturbation to benign images will not make obvious changes. Therefore, for an input image, people add perturbation to the image, then do inferences for both original and perturbed images. If the outputs are different, the input image is regarded as an adversarial example.

Although the aforementioned AE detection method shows promising results on image classification, few works implement it to object detection, which is a more realistic scenario in real world. One significant challenge behind this is choosing the similarity distance metric for outputs of object detectors. Unlike classification problems, of which the output is a 1-D vector that represents the probabilities for each class, the output of object detection is an ensemble structure that contains bounding box location, label index and confidence scores. In order to deal with the difference, the existing works only have simple approaches. Some existed works simply assume there is only one bounding box in the image and fall the problem back to classification problem. In real world, one image often contains multiple objects. Thus, we cannot make the aforementioned assumption in autonomous driving scenario. The other works directly apply mean average precision (mAP) to calculate the distance between the outputs of two images. However, average precision (AP) based metric is designed for calculating similarity distance of multiple images. By using AP, all the bounding boxes are treated as equally weighted. It has to be noticed that for perturbation based AE detection method, even adding perturbation to benign images will causes output differences. The majority difference is the changes of small objects. Due to the reasons discussed above, AP based metric is not suitable for perturbation based AE detection.

In this paper, our goal is to bridge the gap between the current adversarial examples (AE) research and the real-world autonomous systems by providing a unified framework, which allows the AE detection techniques that have been developed for image classification tasks to be easily transferred to the visual perception tasks of autonomous vehicles. To achieve this goal, we first reevaluate the impact of AE in the presence of the end-to-end visual perceptual systems, and then propose a defense leveraging the evaluation results to effectively prevent the threat from AE in self-driving tasks. Code is publicly available at: \href{https://github.com/erbloo/wAP_feature_squeezing}{https://github.com/erbloo/wAP\_feature\_squeezing}.
We make the following contributions:
\begin{itemize}
    \item We empirically validate the threat model of adversarial examples in the self-driving context, and our results show that the object tracking mechanisms in the perception pipeline of autonomous vehicles raise the bar for attack, but are still vulnerable to carefully crafted adversarial examples that reliably deceive object detectors.
    \item We propose a framework for defending autonomous vehicles from the threat of adversarial examples, which extends existing detection mechanisms developed for image classifiers into the object detection domain by (1) designing novel similarity metric to detect inconsistency between dense object detection results of two video frames, and (2) adding temporary information into the detection pipeline to further improve accuracy.
    \item To evaluate our approach, we implements adversarial attacks against state-of-the-art object detectors based on the most effective attacks in the image classification domain. We conduct a large scale evaluation on diverse adversarial targets on different datasets, including Cityscapes~\cite{cordts2016cityscapes} and real-world autonomous driving video datasets~\cite{huang2018apolloscape}. The results shows unprecedentedly high accuracy with x\% FP and x\% FN. 
\end{itemize}
\section{Overview}
\label{sec:overview}

\subsection{Background on Object Detection }

\begin{figure*}[]
    \centering
    \begin{minipage}[b]{0.22\textwidth}
        \includegraphics[width=\textwidth]{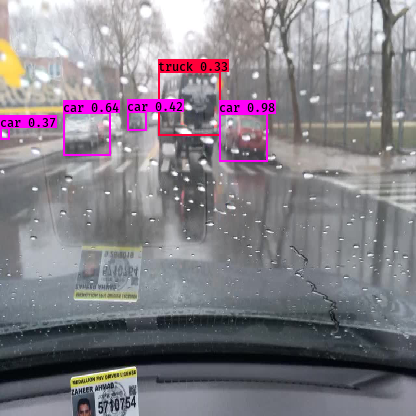}
    \end{minipage}
    \begin{minipage}[b]{0.22\textwidth}
        \includegraphics[width=\textwidth]{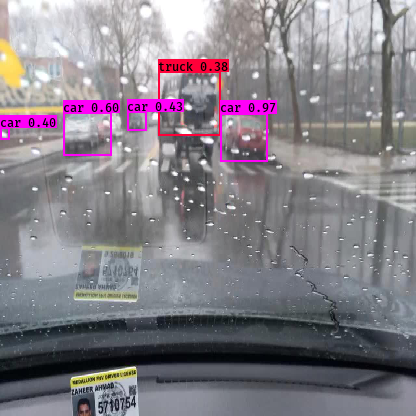}
    \end{minipage}
    \begin{minipage}[b]{0.22\textwidth}
        \includegraphics[width=\textwidth]{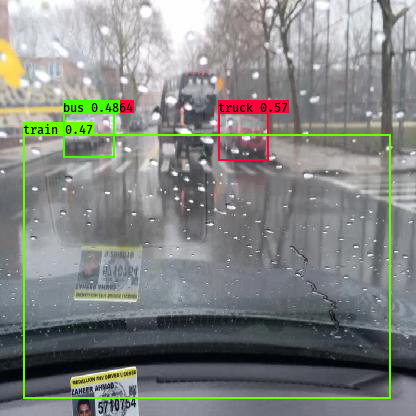}
    \end{minipage}
    \begin{minipage}[b]{0.22\textwidth}
        \includegraphics[width=\textwidth]{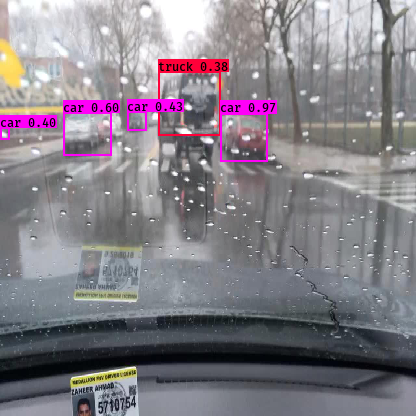}
    \end{minipage}

    \begin{minipage}[b]{0.22\textwidth}
        \includegraphics[width=\textwidth]{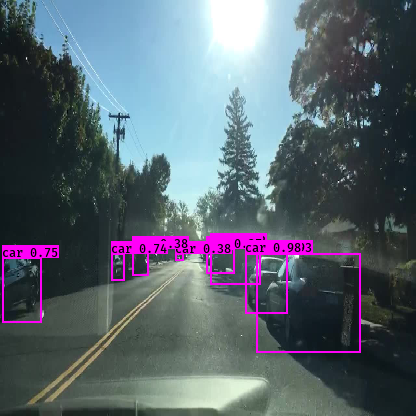}
    \end{minipage}
    \begin{minipage}[b]{0.22\textwidth}
        \includegraphics[width=\textwidth]{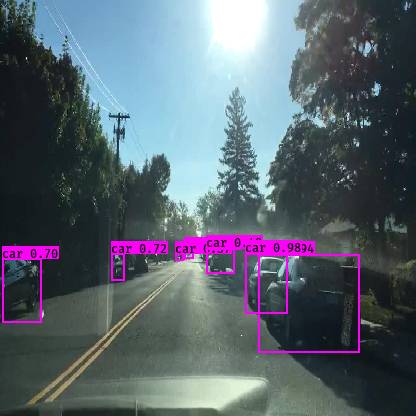}
    \end{minipage}
    \begin{minipage}[b]{0.22\textwidth}
        \includegraphics[width=\textwidth]{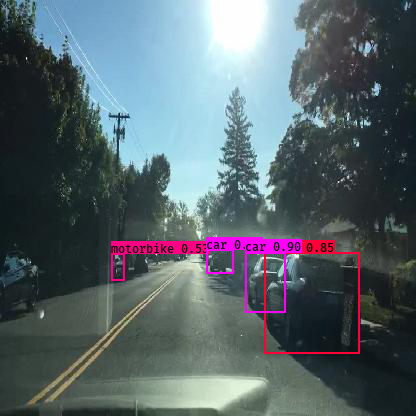}
    \end{minipage}
    \begin{minipage}[b]{0.22\textwidth}
        \includegraphics[width=\textwidth]{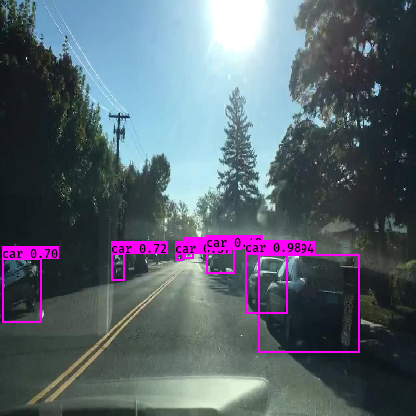}
    \end{minipage}

    \begin{minipage}[b]{0.22\textwidth}
        \includegraphics[width=\textwidth]{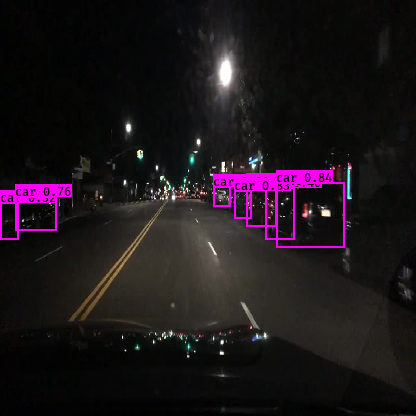}
        \subcaption{Benign.}
    \end{minipage}
    \begin{minipage}[b]{0.22\textwidth}
        \includegraphics[width=\textwidth]{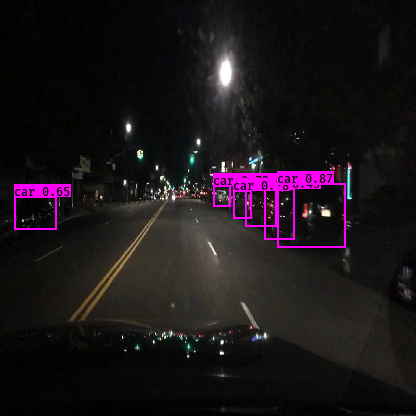}
        \subcaption{Benign Squeeze.}
    \end{minipage}
    \begin{minipage}[b]{0.22\textwidth}
        \includegraphics[width=\textwidth]{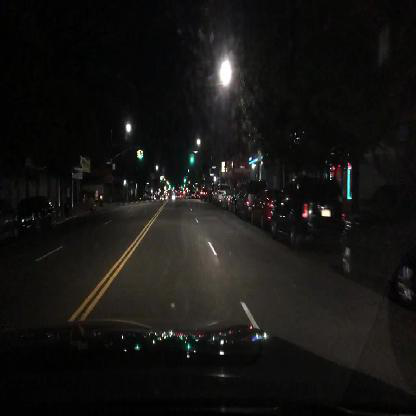}
        \subcaption{Adversarial Example.}
    \end{minipage}
    \begin{minipage}[b]{0.22\textwidth}
        \includegraphics[width=\textwidth]{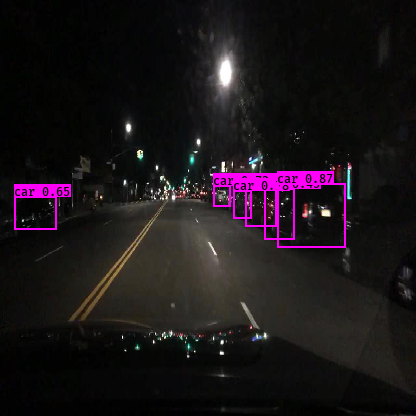}
        \subcaption{AE Squeeze.}
    \end{minipage}
    \caption{
    Sample of feature squeezing for benign and adversarial samples.
    }
    \label{fig:fpfn_roc}
\end{figure*}

Deep convolutional neural networks have been applied to object detection successfully\cite{ylu04}\cite{ylu07}\cite{ylu08}. Currently, one of the most popular object detection pipeline ~\cite{fasterrcnn}~\cite{maskrcnn} uses two stages processing: the first stage generates a number of proposals of different scales and positions, and the second stage classifies the proposals into foreground classes and performing post-processing such as non-maximal suppression (NMS). On the other hand, one stage based method ~\cite{yolov3}~\cite{SSD}~\cite{retinanet} recently made remarkable improvements. One stage detection methods generate dense proposal bounding boxes from intermediate feature maps of backbones. Meanwhile, for each bounding box, regression and classification are performed simultaneously to obtain the final outputs.

\subsection{Adversarial Examples}
Carlini and Wagner (C\&W) attack proposed different gradient based attacks using different $L_p$ norms, namely $CW_2$, $CW_{\infty}$, $CW_0$ attacks. It has been recognized as the most effective attacks against DNNs, and in Madry \etal~\cite{madry2017towards}, it has been shown that a C\&W attack is a universal adversary among all first-order attack methods. Their analysis and evidence from experiments all suggest any adversarial attack method that only incorporates gradients of the loss function w.r.t. the input cannot do significantly better than it.

\textbf{Existing Defense.} Defense mechanisms aims at preventing AE by hardening the DNN models, and are summarized by Papernot \etal~\cite{papernot2016towards} into two broad categories: \emph{adversarial training} and \emph{gradient masking}. Goodfellow \etal first introduced the idea of adversarial training~\cite{kurakin2016adversarial}, which try to integrate existing adversarial existing AE generation methods into the training process so that the trained model can easily defend such attacks~\cite{madry2017towards, tramer2017ensemble}. Recent work have also extended the idea of adversarial training to obtain certified robustness against certain attacks by introducing either abstract interpretation or perturbations into the training process~\cite{mirman2018differentiable, li2018second, kolter2017provable, wong2018scaling}. However, it requires prior knowledge of possible attacks and the robustness of adversarially trained model usually overfits to the choice of norms~\cite{carlini2017towards}.

The idea of gradient masking is to enhance the training process by training the model with small gradients so that the model will not be sensitive to small changes in the input~\cite{gu2014towards}. However, it was reported such models can be broken by advanced attacks. Papernot \etal~\cite{papernot2016towards} concluded that controlling gradient information in training has limited effects in defending adversarial attacks.

\textbf{Existing Detection.} Compared with those defense mechanisms, adversarial example detection choose to prevent the model from making predictions based on suspicious input instead of hardening the model itself. The existing state-of-the-art AE detection works can be classified into two major categories.

\subsubsection{Metric based detection} Researchers have proposed to perform statistical measurement of the inputs (and activation values) to detect adversarial samples\cite{ylu10}\cite{ylu11}\cite{ylu12}\cite{ylu13}. Feinman \etal~\cite{feinman2017detecting} proposed to use the \emph{Kernel Density} estimation (KD) and Bayesian Uncertainty (BU) to identify adversarial subspace to separate benign inputs and adversarial samples. Carlini and Wagner~\cite{carlini2017adversarial} showed this method can be bypassed but also commented this method to be one of the promising directions. Inspired by ideas from the anomaly detection community, Ma \etal~\cite{ma2018characterizing} recently proposed to use a measurement called Local Intrinsic Dimensionality (LID). For a given sample input, this method estimates a LID value which assesses its space-filling capability of the region surrounding the sample by calculating the distance distribution of the sample and a number of neighbors for individual layers. The authors empirically show that adversarial samples tend to have large LID values. Their results demonstrate that LID outperforms BU and KD in AE detection and currently represents the state-of-the-art for this kind of detectors. However, a key challenge of these techniques is how to define a high-quality statistical metric which can clearly tell the difference between clean samples and adversarial samples. Lu \etal~\cite{lu2018limitation} have shown that LID is sensitive to the confidence parameters deployed by an attack and vulnerable to transferred adversarial samples.

\subsubsection{Prediction inconsistency based detection} Many other works are based on the prediction inconsistency~\cite{xu2017feature, dhillon2018stochastic, guo2017countering}. These works are built based a empirically validated hypothesis that \emph{the robustness of DNNs to local changes (e.g., squeezing, scale, position) does not generalize to the perturbations added by adversarial examples.} Thus they propose different transformation methods $T(\cdot)$ either on input samples expecting the transformed sample will produce different prediction results as the original sample if it is adversary~\cite{xu2017feature, guo2017countering, meng2017magnet, luo2015foveation, xiao2018characterizing}, or on the model expecting that the transferred model will make predictions different than the original model if the input sample is adversarially perturbed~\cite{tao2018attacks}. For example, transformation on inputs using foveation~\cite{luo2015foveation}, smoothing~\cite{xu2017feature}, scaling~\cite{xiao2018characterizing}, denoising~\cite{meng2017magnet}, and transformation on model by strengthening interpretability~\cite{tao2018attacks} have all been proposed. Our  framework allows all of these auxiliary transformations that have been demonstrated effective in unveiling adversarial examples in image classification and face recognition to be easily transferred to the object detection domain where new challenges present (\S\ref{sec:design}) without sacrificing effectiveness.

\subsection{Notions}
\begin{itemize}
 \item Testing image input $x$, model predicts $f(x)$.
 \item Adversary example $x'=x+\varepsilon$, $\varepsilon$ is the perturbation.
\end{itemize}
Detecting adversarial examples:
\begin{itemize}
    \item Define a distance $D(x,y)=||x-y||_R$
    \item Find a transformation $T(\cdot)$
    \item For normal testing image $D(f(x),f(T(x))) < \mu$
    \item For adversarial image $D(f(x'),f(T(x')) \gg \mu$
\end{itemize} 
\section{Re-estimation of Adversarial Threat}

Autonomous systems employ object tracking algorithms to post-process the object detection results to mitigate the variance and uncertainty of the sensor estimate, before using them to produce control decisions. Taking an initial set of object detections (set of bounding boxes and labels), an object tracking algorithm creates a unique ID for each of the initial detections  and tracks each of the objects as they move around frames in a video.

\begin{figure}
\centering
\includegraphics[width=0.9\columnwidth]{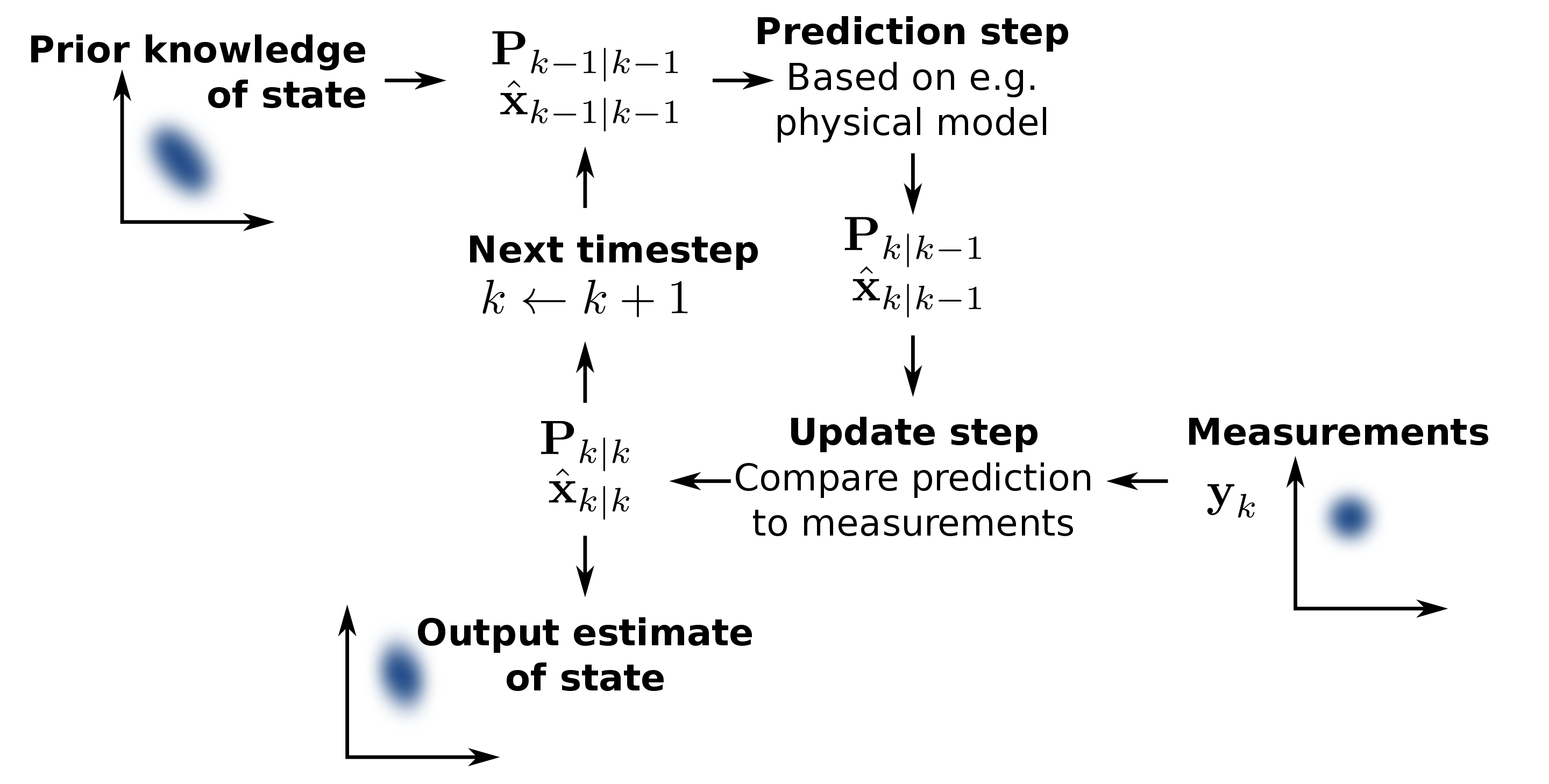}
\caption{Illustrated work flow of Kalman filter.}
\label{fig:kalman}
\end{figure}

An ideal object tracking algorithm is able to handle when the tracked object "disappears" or moves outside the boundaries of the video frame, and is able to pick up objects it has "lost" in between frames. Also it is robust to occlusion. Kalman filter~\cite{brown1992introduction}, also known as linear quadratic estimation (LQE), is the most commonly used object tracking algorithm, and is employed by in the popular open source self-driving platforms including Apollo~\cite{apollo} and Autoware~\cite{kato2015open}. Shown in Figure~\ref{fig:kalman}, a Kalman filter first produces estimates of the current state variables, along with their uncertainties, and once the outcome of the next measurement (necessarily corrupted with some amount of error, including random noise) is observed, these estimates are updated using a weighted average, with more weight being given to estimates with higher certainty. The algorithm is recursive. It can run in real time, using only the present input measurements and the previously calculated state and its uncertainty matrix; no additional past information is required. Since Kalman filter is designed and has been proven effective in tolerating for missing frames and occasional inaccurate object detection results~\cite{sun2004multi}, it's nature to review the hypothesis that adversarial examples that are not strong enough to reliably deceive object detectors have only limited effects on the perception results.

Recent works~\cite{jia2019tracking} study adversarial machine learning attacks considering the practical and complete visual perception pipeline in autonomous driving. Both object detection and object tracking are taken into consideration. The study shows that because of the existence of Kalman filter based tracking system, only temporal adversarial attacks that successfully attack several frames are able to fool the whole perception pipeline in autonomous driving. However, current physical adversary attacking technologies are not robust enough to generate high success rate AEs in consecutive frames. Thus, by aforementioned research, we assume physical adversary examples that are not robust enough to consistently fool object detectors have only limited effects on perception results due to Kalman filter on autonomous vehicles.

The threat re\-estimation results suggest that though Kalman filter raises the bar for adversarial attack, strong and robust adversarial examples can still fool the detector and cause wrong perception results.


\section{Design}
\label{sec:design}

\begin{figure*}[h]
\centering
\includegraphics[width=0.9\textwidth]{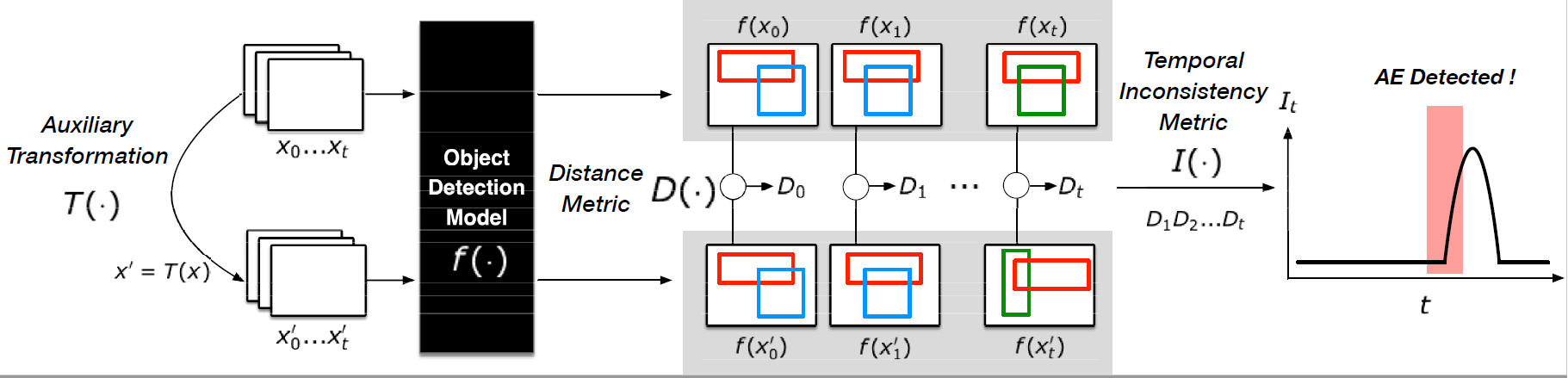}
\caption{Proposed framework for detecting adversarial examples that can reliably deceive object detectors. The model is evaluated on both the original frames and the pre-processed frames after a chosen auxiliary transformation $T(\cdot)$. A distance $D\cdot$ is calculated on the object detection results of each pair of frames. The distances on consecutive frames used to further obtain a temporal inconsistency metric $I(\cdot)$, and the adversarial attack can be detected on the fly by monitoring the variance of $I$ over time.}
\label{fig:design}
\end{figure*}

\subsection{Overview}
Figure~\ref{fig:design} shows an overview of our AE detection framework. First, for each frame $x_t$ coming from the camera at time $t$, the system obtains a transformed input $x'_t$ using an auxiliary transformation $T(x)$. It is based on a hypothesis that the robustness of DNNs to local changes (e.g., squeezing, scale, position) does not generalize to the perturbations added by adversarial examples, which have been validated by previous AE detection works~\cite{luo2015foveation, xu2017feature, meng2017magnet}. Thus if $x$ is an adversarial example, it is highly likely that the object detection result $f(x)$ is very different from $f(x')$.

However, there is no available metric to describe and quantify the similarity of two object detection results. Previous adversarial detections on image classification tasks simply need to compare the predicted classes and their confidences for differences, while in object detection, we need to deal with dense and sometimes overlapped bounding boxes, and it's non-trivial to design a metric that balances precision and recall well. Mean average precision (mAP) is the most commonly used metric to evaluate the performance of an object detector, but we will show later that the false positives can be prohibitively high if using mAP as the distance metric, since it assigns all equal weights to detected objects regardless of their sizes and classes. Also calculating mAP between two frames instead of a set of testing images provides no statistical significance.

Thus we design a novel distance metric $D(x,y)$ to precisely describe the differences between two object detection results (detailed in \S\ref{sec:distance_metric}).

At last by monitoring the variance of a temporal inconsistency metric $T(D_t)$, adversarial examples can be detected in real time, and mitigation actions, such as requiring human control, or rolling back to the predictions by Kalman filter could be taken in time.

\subsection{Frame-based Distance Metric $D(x,y)$ }
\label{sec:distance_metric}

\begin{algorithm}[h]
 \KwData{ground truth: \{bounding boxes, confidence score, object class\},
         prediction: \{bounding boxes, confidence score, object class\}}
 \KwResult{Detection distance }
 tp = [] \;
 fn = [] \;
 \For{$i\gets0$ \KwTo $N_{gt}$}{
    \For{$j\gets0$ \KwTo $N_{pd}$}{
        $IoU_{ij} \gets getIoU(gt_i.bbox, pd_j.bbox)$\;
        \If{$IoU_{ij} > MIN_{OVERLAP}$ \&\& $gt_i.cl == pd_j.cl$}{
            tp append $(i, j)$ \;
        }
    }
    fn append $i$ \;
 }
 fp = \{0, 1, ..., $N_{pd}$\} \;
 \For{$i\gets0$ \KwTo $N_{tp}$}{
    del fp[tp[i,1]] \;
 }
 $\mathcal{D}_{tp} = \mathcal{F} (\frac{\sum_{(i, j)\in tp} \mathcal{DA}(gt_i.bbox, pd_j.bbox) }{A_{pd} + A_{gt}}) + \gamma_{cs}\sum_{(i, j)\in tp}\mid{C_{gt_i}-C_{pd_j}}\mid $ \;
 $\mathcal{D}_{fp} = \mathcal{F} (\frac{\sum_{i\in fp} A_{pd_i} }{A_{pd}}) + \gamma_{cs}\sum_{i\in fp} {C_{pd_i}}$ \;
 $\mathcal{D}_{fn} = \mathcal{F} (\frac{\sum_{i\in fn} A_{gt_i} }{A_{gt}}) + \gamma_{cs}\sum_{i\in gt} {C_{gt_i}}$ \;
 $D(x,y) = \frac{\alpha_{tp}\mathcal{D}_{tp} + \alpha_{fp}\mathcal{D}_{fp} + \alpha_{fn}\mathcal{D}_{fn}}{\alpha_{tp}+\alpha_{fp}+\alpha_{fn}}$

 \caption{Frame-based Distance Metric $D(x,y)$}
\end{algorithm}

We propose a frame-based distance metric given detection results of two images $x$ and $y$. From output of image detector$I=\{I_0, I_1,...,I_N\}, I\in\{x,y\}$, bounding boxes positions($bbox$), bounding boxes confident scores($cs$) and bounding boxes predicted classes($cl$) are provided for each input image. We refer detection result of image $x$ as ground truth and detection result of image $y$ as prediction. Let $tp$ be the set of pairs of true positive bounding boxes indexes, $fp$ be the set of indexes of false positive prediction bounding boxes and $fn$ be the set of indexes of false negative ground truth bounding boxes. Based on Intersection over Union(IoU) values of ground truth bounding boxes, prediction bounding boxes and given minimal overlapping threshold($MIN_{OVERLAP}$), we classify bounding boxes into true positive set, false positive set and false negative set. Distance is composed of two parts: area based distance and confidence score based distance. For bounding boxes pairs in true positive set, difference area($\mathcal{DA}(A, B) = (A-B)\cup(B-A)$) is calculated and divided by sum of areas of both ground truth and prediction bounding boxes. For false positive\slash false negative set, bounding boxes areas are calculated and divided by sum of areas of prediction\slash ground truth bounding boxes areas. All the area based values are fed into weighed function $\mathcal{F}(x)=\frac{x}{x+a}$ to make small bounding boxes contribute less. For confident scores part, we obtain difference of confident scores of true positive samples, confident scores of false positive samples and false negative samples. All the confident score based values are multiplied by a constant weight parameter($\gamma_{cs}$). $\mathcal{D}_{tp}$, $\mathcal{D}_{fp}$ and $\mathcal{D}_{fn}$ are weighted summed to get the final distance$D(x,y)$.

\begin{figure*}[h!]
    \centering
    \begin{minipage}[b]{0.23\textwidth}
        \includegraphics[width=\textwidth]{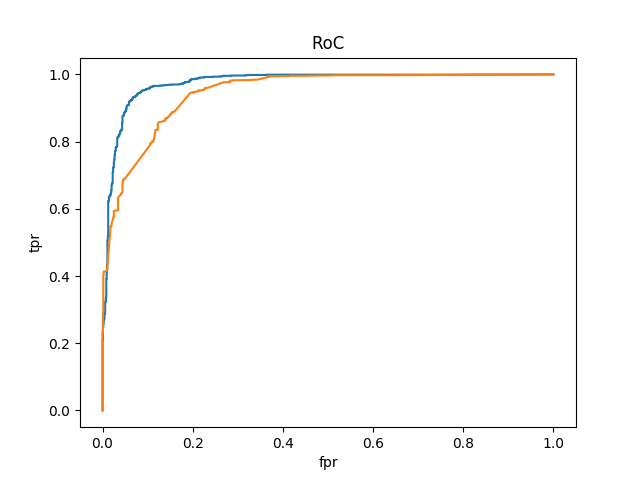}
        \subcaption{C\&W Bit4 ROC.}
    \end{minipage}
    \begin{minipage}[b]{0.23\textwidth}
        \includegraphics[width=\textwidth]{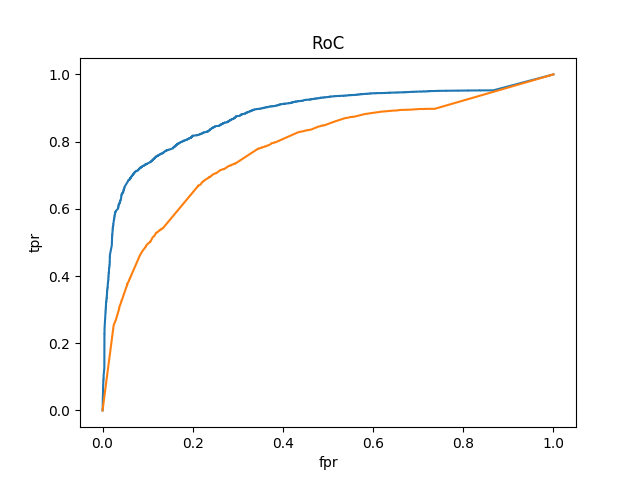}
        \subcaption{C\&W Bit5 ROC.}
    \end{minipage}
    \begin{minipage}[b]{0.23\textwidth}
        \includegraphics[width=\textwidth]{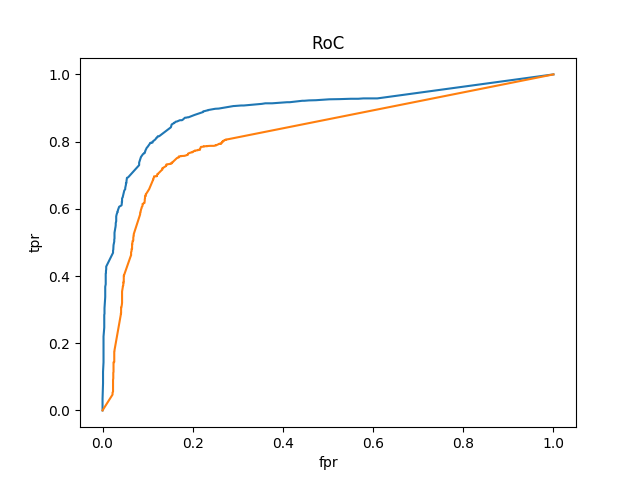}
        \subcaption{C\&W Bit6 ROC.}
    \end{minipage}
    \begin{minipage}[b]{0.23\textwidth}
        \includegraphics[width=\textwidth]{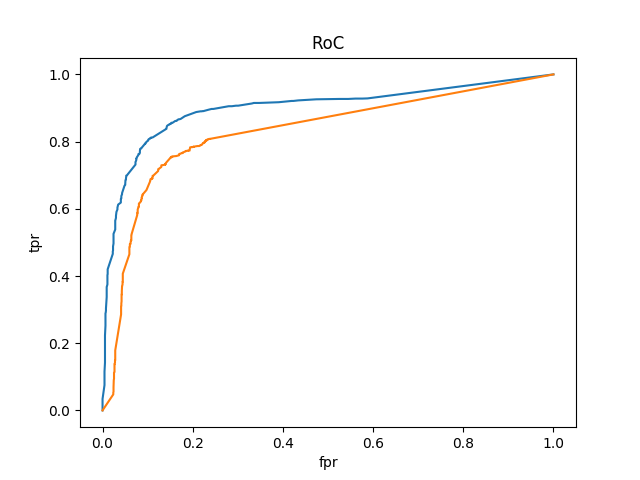}
        \subcaption{C\&W Bit7 ROC.}
    \end{minipage}

    \begin{minipage}[b]{0.23\textwidth}
        \includegraphics[width=\textwidth]{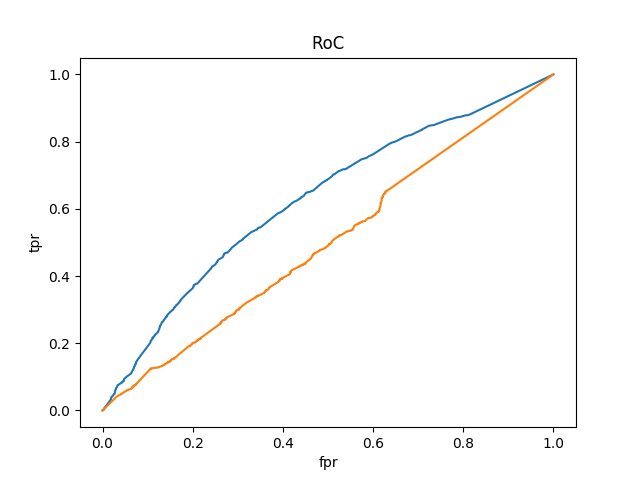}
        \subcaption{Gaussian Noise Bit4 ROC.}
    \end{minipage}
    \begin{minipage}[b]{0.23\textwidth}
        \includegraphics[width=\textwidth]{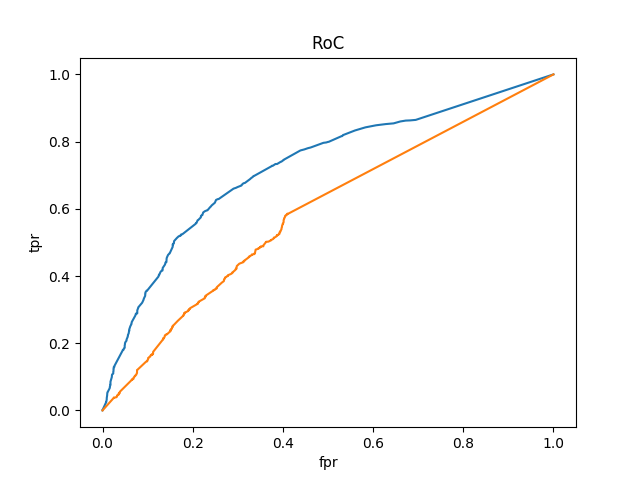}
        \subcaption{Gaussian Noise Bit5 ROC.}
    \end{minipage}
    \begin{minipage}[b]{0.23\textwidth}
        \includegraphics[width=\textwidth]{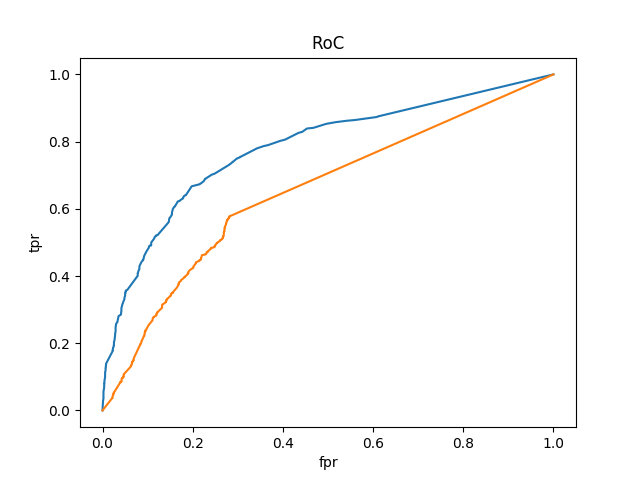}
        \subcaption{Gaussian Noise Bit6 ROC.}
    \end{minipage}
    \begin{minipage}[b]{0.23\textwidth}
        \includegraphics[width=\textwidth]{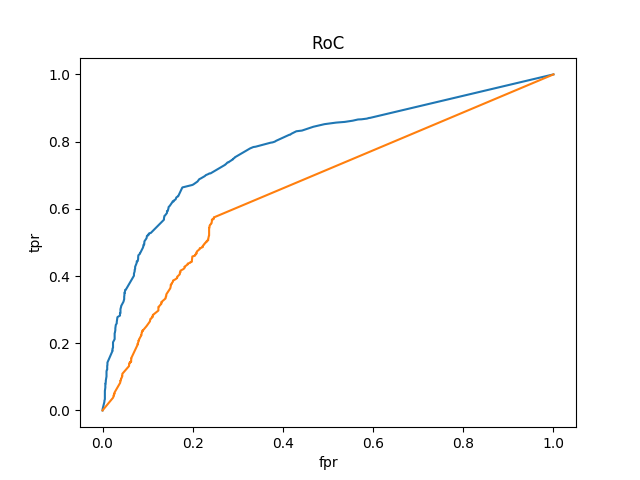}
        \subcaption{Gaussian Noise Bit7 ROC.}
    \end{minipage}

    \begin{minipage}[b]{0.23\textwidth}
        \includegraphics[width=\textwidth]{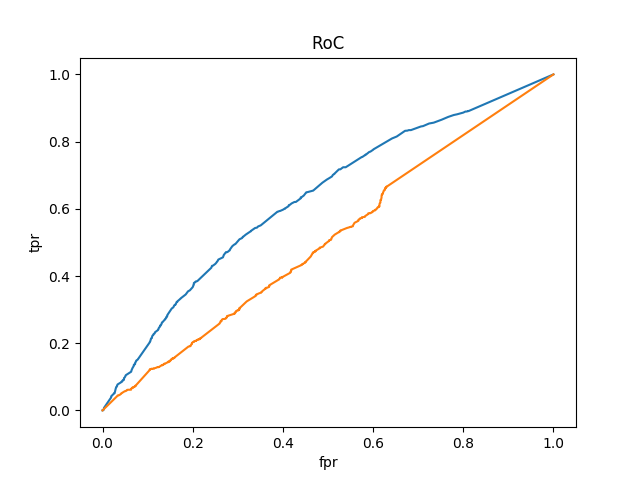}
        \subcaption{Brightness Bit4 ROC.}
    \end{minipage}
    \begin{minipage}[b]{0.23\textwidth}
        \includegraphics[width=\textwidth]{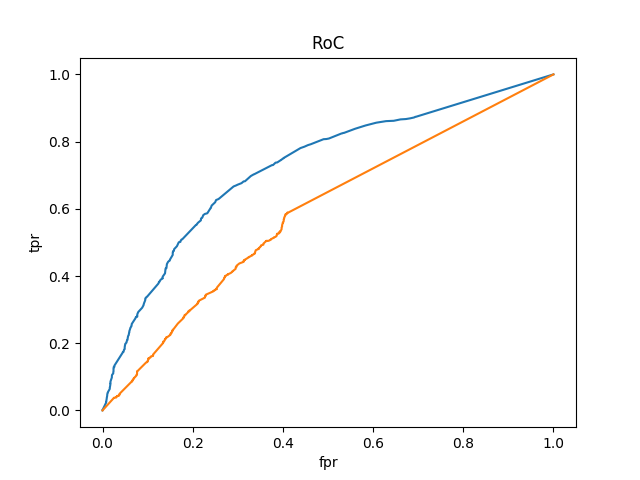}
        \subcaption{Brightness Bit5 ROC.}
    \end{minipage}
    \begin{minipage}[b]{0.23\textwidth}
        \includegraphics[width=\textwidth]{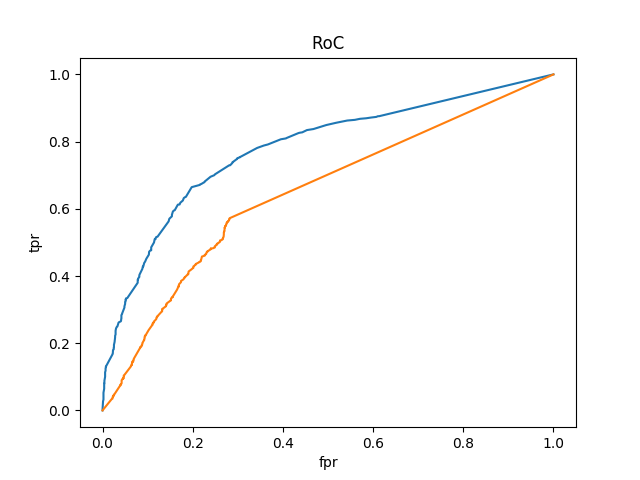}
        \subcaption{Brightness Bit6 ROC.}
    \end{minipage}
    \begin{minipage}[b]{0.23\textwidth}
        \includegraphics[width=\textwidth]{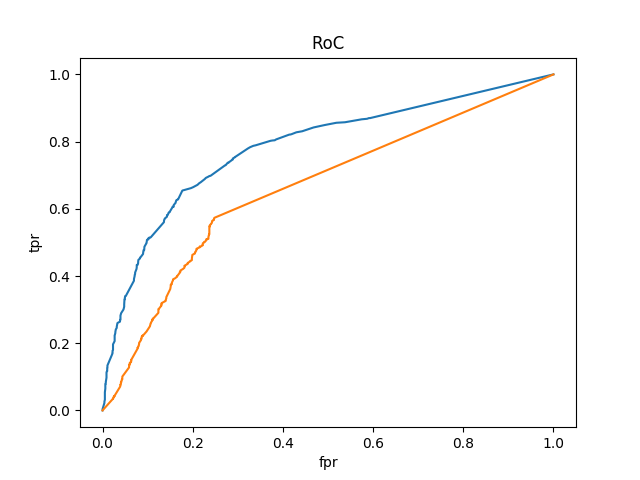}
        \subcaption{Brightness Bit7 ROC.}
    \end{minipage}
    \caption{
    Results of FP-FN and ROC curves for mAP and the proposed wAP metric.
    }
    \label{fig:fpfn_roc}
\end{figure*}

\subsection{Temporal Consistency}
\label{ssec:temporal}
In perception system, a tracker will be deleted if no objects is associated with for a duration of R frames, or called a reserved age ($R$). It prevents the tracks from being accidentally deleted due to infrequent false negatives of object detectors.~\cite{jia2019tracking} Thus, for detection algorithm for AE, only attacks that last longer than reserved age should be regarded as a valid adversarial example. Inspired by this fact, we propose a temporal detection algorithm $T(D_t)$, which can be explained as

\begin{equation}
\begin{aligned}
&y=T(D_0 ... D_t)=\prod_{i=0}^{t}(sgn(D_i-\theta)) \\
\end{aligned}
\label{eq:Temporal}
\end{equation}

Where $sgn$ is signum function and $\theta$ is the threshold for distance metric $D(\cdot)$.

\section{Experimental Results}
\label{sec:experiment}

\subsection{Experimental Setup}
Original testing data set comes from BDD10k~\cite{bdd} and Cityscapes~\cite{cordts2016cityscapes} datasets. We implement both white-box and black-box attacks and insert adversarial frames into the videos to build our adversarial evaluation set. For white-box attack, we implement $CW_{inf}$ attack that is the state of the art white-box attacking methodology. For black-box attack, we implement Gaussian Noise and Brightness attacks. We choose best performance feature squeezing methods: bit-wise squeeze. For generating object detection outputs, we use Yolo-v3~\cite{yolov3}, which is widely implemented in autonomous driving perception system.

\subsection{Evaluation on Single Frame Adversarial Example Detection}
Single frame adversarial examples results are shown in Figure~\ref{fig:fpfn_roc} and Table~\ref{tab:det_table}.

Figure~\ref{fig:fpfn_roc} shows ROC curves for mAP and proposed wAP methods. C\&W, Gaussian Noise and Brightness attacks are performed and Bit4, Bit5, Bit6 and Bit7 are implemented as squeezing methods.

\begin{figure*}[]
\centering
\includegraphics[width=0.9\textwidth]{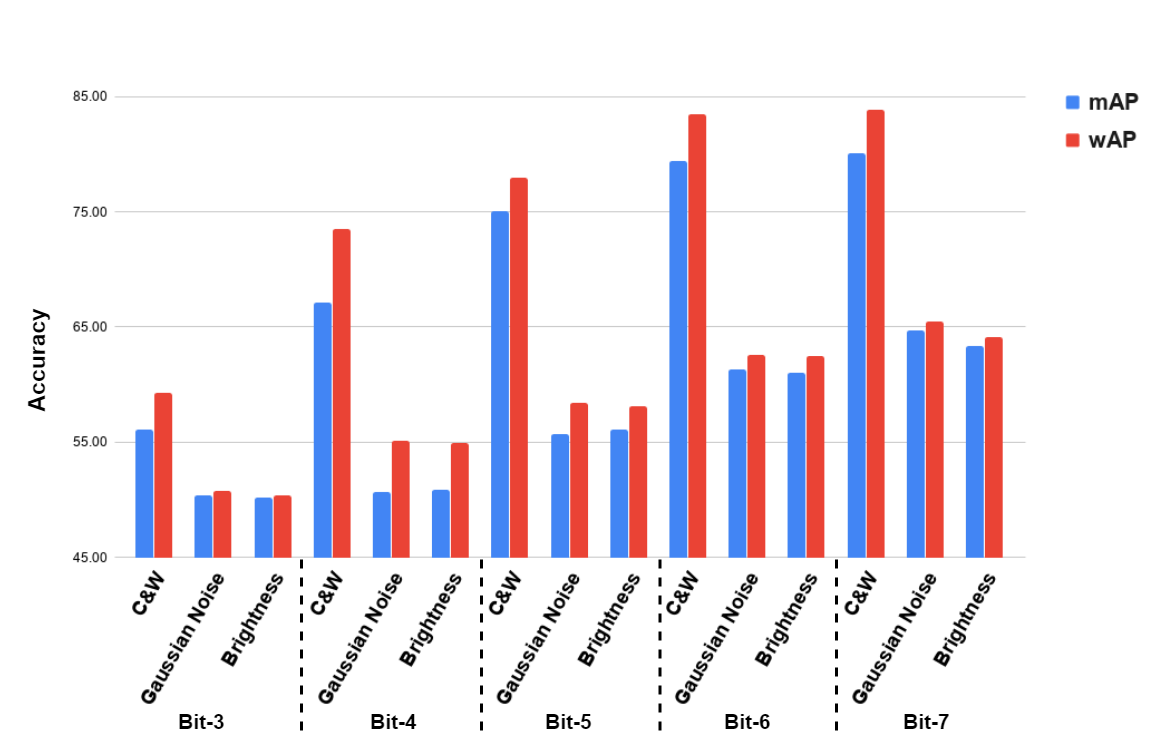}
\caption{Overall results comparison for proposed method and mAP. The accuracies are calculated by averaging the results of Bdd10k and Cityscapes.}
\label{fig:wAP_mAP}
\end{figure*}

Table~\ref{tab:det_table} presents the detailed accuracy results of single frame experiments using testing set of BDD10k and CityScapes. In the table, C\&W, Gaussian Noise and Brightness attacks are implemented and Bit(3, 4, 5, 6, 7) squeezes are used as AE detection algorithm. We can see that our proposed method outperformed the mAP based algorithm on both Bdd10k and CityScapes datasets. For 27 over 30  experiments using black-box and white-box attacks, our proposed method shows higher detection accuracies. Bit7 shows better results comparing to other bits squeezes. For instance, the proposed method reports 83.77\% and 83.88\% accuracy on C\&W-Bit7 experiment, respectively. The accuracies are 3.71\% and 3.77\% more that that of mAP metric. Figure~\ref{fig:wAP_mAP} shows the statistical results. The average accuracy of two datasets are shown in the figure. We can see that the proposed wAP algorithm outperforms mAP baselines in all experiments.

\begin{table*}[]
\centering
\begin{tabular}{l|l|ll|ll}
\hline
Accuracy(\%)       &              & \multicolumn{2}{l|}{Bdd10k} & \multicolumn{2}{l}{Cityscapes}  \\ \hline
Attack mtd.    & Squeeze mtd. & mAP      & Proposed         & mAP            & Proposed       \\ \hline
C\&W           & bit3         & 56.20    & \textbf{59.24}   & 56.02          & \textbf{59.31} \\
               & bit4         & 67.23    & \textbf{73.46}   & 67.03          & \textbf{73.55} \\
               & bit5         & 75.41    & \textbf{75.68}   & 74.72          & \textbf{80.20} \\
               & bit6         & 79.49    & \textbf{83.41}   & 79.37          & \textbf{83.50} \\
               & bit7         & 80.06    & \textbf{83.77}   & 80.11          & \textbf{83.88} \\ \hline
Gaussian Noise & bit3         & 50.34    & \textbf{50.85}   & 50.45          & \textbf{50.67} \\
               & bit4         & 51.14    & \textbf{59.47}   & 50.22          & \textbf{50.87} \\
               & bit5         & 58.68    & \textbf{64.26}   & 52.64          & 52.64          \\
               & bit6         & 64.79    & \textbf{67.17}   & 57.93          & \textbf{58.04} \\
               & bit7         & 66.36    & \textbf{67.47}   & 63.10          & \textbf{63.54} \\ \hline
Brightness     & bit3         & 50.21    & \textbf{50.75}   & \textbf{50.23} & 50.00          \\
               & bit4         & 51.78    & \textbf{59.48}   & 50.00          & \textbf{50.34} \\
               & bit5         & 58.84    & \textbf{63.03}   & 53.27          & 53.27          \\
               & bit6         & 64.51    & \textbf{65.93}   & 57.55          & \textbf{59.12} \\
               & bit7         & 66.25    & \textbf{66.69}   & 60.38          & \textbf{61.51} \\ \hline
\end{tabular}
\caption{
    Results of detection accuracies using C\&W, Gaussian Noise and Brightness attacks to different feature squeezing methods.
}
\label{tab:det_table}
\end{table*}

\subsection{Evaluation on Temporal Adversarial Example Detection}
As discussed in ~\cite{jia2019tracking}, the state of the art temporal attack is able to successfully attack tracking system in 3 frames. Thus, we set temporal frames $i=3$. Figure~\ref{fig:temporal} shows the accuracy results that compare the performance between single frame and temporal experiments. Bit7 is used as squeezing method, since it shows the best performance in Table~\ref{tab:det_table}. We can see that both temporal based AE detection results from mAP and wAP have better results than single frame based results. This shows that the algorithm~\ref{ssec:temporal} improves the detection performance. It should be noticed that the proposed temporal wAP algorithm achieves the best accuracy result for all the attacking methods.

\begin{table}[]
\begin{tabular}{c|c|ccc}
\hline
\multicolumn{2}{c|}{Detection}                 & C\&W           & \begin{tabular}[c]{@{}c@{}}Gaussian\\ Noise\end{tabular} & Brightness \\ \hline
\multirow{2}{*}{Single Frame} & mAP            & 67.61          & 62.85                                                    & 60.14      \\
                              & wAP (Proposed) & 72.05          & 69.50                                                    & 68.12      \\ \hline
\multirow{2}{*}{Temporal}     & mAP            & 80.82          & 73.22                                                    & 71.11      \\
                              & wAP (Proposed) & \textbf{85.37} & \textbf{74.61}                                                    & \textbf{73.79}      \\ \hline
\end{tabular}
\caption{Results comparison for proposed method and mAP using both single frame and temporal detection scenarios.}
\label{fig:temporal}
\end{table}

\section{Discussion}
\label{sec:discussion}

\subsection{Applicability to other Perception Tasks}
As discussed in \cite{xu2017feature}, the feature-squeezing approach could be used in many domains where deep learning is used. Additionally, the proposed wAP metric is calculated based on IoU. Thus, any overlapping based perception tasks such as object detection, semantic segmentation and text detection \& recognition are fitful to the proposed detecting algorithm. As for proposed temporal algorithm, whatever temporal and tracking systems that based tasks such as video and audio processing can be applied. 

\subsection{Adaptive Attack}
Comparing to mAP baseline distance metric, our proposed wAP metric is the optimized version, which focuses on single image instead of the whole dataset and introduces weights to bounding boxes. Thus, for different sorts of attacks, even any of them that is able to effectively attacking the detecting pipeline, the proposed metric may still perform better than mAP. 
\section{Conclusion}
\label{sec:conclusion}

In this paper, we propose a \emph{weighted-AP} (wAP) distance metric and temporal optimization method to improve the detecting of adversarial example for object detection in autonomous driving perception system. Specifically, our proposed metric focuses on single image bounding boxes and is applied on sequential frames to fit tracking system. Compared to existing single frame mAP detecting method, the results on BDD and CityScapes show that our proposed method performs better on detecting adversarial examples for autonomous driving perception system.
One intuition behind the temporal wAP is that fooling single frame is not enough to successfully attack the perception system. This is because tracking system is able to make predictions when bounding boxes of objects miss for a small number of frames. Moreover, practical object detection outputs contain dense bounding boxes that cannot easily obtain a scalar distance between two output results and mAP is a metric widely used for dataset based evaluation but not for calculating distance for two single images. Evaluation on different autonomous driving datasets show that this proposed pipeline greatly enhance detecting performance by a large margin compared to the baseline method. We hope that our proposed distance metric and temporal solution can make research that are related to detecting of object detection more realistic. Code is publicly available at: \href{https://github.com/erbloo/wAP_feature_squeezing}{https://github.com/erbloo/wAP\_feature\_squeezing}.

\medskip

\clearpage
\setcounter{page}{1}
\bibliographystyle{IEEEtran}
\bibliography{ref}

\end{document}